\documentclass[11pt, a4paper, logo, copyright, nonumbering]{map}

\usepackage{pifont}
\PassOptionsToPackage{table,xcdraw}{xcolor}
\usepackage{geometry}
\usepackage{multirow}
\usepackage[most]{tcolorbox}
\usepackage[justification=justified]{caption}
\usepackage{subcaption} % 需要在导言区引入
\usepackage{multicol}
\definecolor{cvprblue}{rgb}{0.21,0.49,0.74}

\usepackage{natbib}

\usepackage{CJKutf8}

\usepackage{xargs}  

\usepackage{todonotes}  

\usepackage{multirow}

\usepackage{cleveref}

\usepackage{amsmath}
\usepackage{dsfont}

\usepackage{subcaption}

\usepackage{svg}
\PassOptionsToPackage{most}{tcolorbox}  % 在主文档的开头传递选项
\usepackage{tcolorbox}  % 然后加载包
\usepackage{fontawesome}
\usepackage{xcolor}

\usepackage[utf8]{inputenc}
\usepackage{booktabs}
\usepackage{graphicx}
\usepackage{array}
\usepackage{tabularx}
\usepackage{xcolor,colortbl}  % 引入xcolor包
\definecolor{boxcolor}{HTML}{d92523} % 框的颜色 #dea3a2
\definecolor{bulbcolor}{HTML}{e3b87f} % 灯泡的颜色 #e3b87f

\hypersetup{
    colorlinks=true,            %链接颜色
    linkcolor=blue,             %内部链接
    filecolor=magenta,          %本地文档
    urlcolor=cyan,              %网址链接
    citecolor=cvprblue,             % 引用颜色
    pdftitle={Overleaf Example},
    pdfpagemode=FullScreen,
    }
 
\renewcommand{\today}{}
\newcommandx{\info}[2][1=]{\todo[linecolor=red,backgroundcolor=red!25,bordercolor=red,#1]{#2}}

\title{\centering  ScaleLong: A Multi-Timescale Benchmark for Long Video Understanding}

\author{
\textbf{M-A-P} \\
ByteDance Inc.
}

\begin{abstract}
Although long-video understanding demands that models capture hierarchical temporal information—from clip (seconds) and shot (tens of seconds) to event (minutes) and story (hours)—existing benchmarks either neglect this multi-scale design or scatter scale-specific questions across different videos, preventing direct comparison of model performance across timescales on the same content. To address this, we introduce ScaleLong, the first benchmark to disentangle these factors by embedding questions targeting four hierarchical timescales\textemdash clip (seconds), shot (tens of seconds), event (minutes), and story (hours)\textemdash all within the same video content. This within-content multi-timescale questioning design enables direct comparison of model performance across timescales on identical videos. ScaleLong features 269 long videos (avg. 86 min) from 5 main categories and 36 sub-categories, with 4–8 carefully designed questions, with at least one question targeting each timescale. Evaluating 23 MLLMs reveals a distinct U-shaped performance trend: higher accuracy at the shortest (clip) and longest (story) timescales, with a dip at intermediate levels. Furthermore, ablation studies demonstrate that increased visual token capacity consistently enhances reasoning across all timescales. ScaleLong offers a crucial fine-grained, multi-timescale benchmark for advancing MLLM capabilities in long-video understanding. The code and dataset are available at \url{https://github.com/multimodal-art-projection/ScaleLong}

\end{abstract}

\begin{document}
\begin{CJK*}{UTF8}{gbsn}

\maketitle

\begin{figure}[htbp]
    \centering
    \includegraphics[width=4.5in, keepaspectratio]{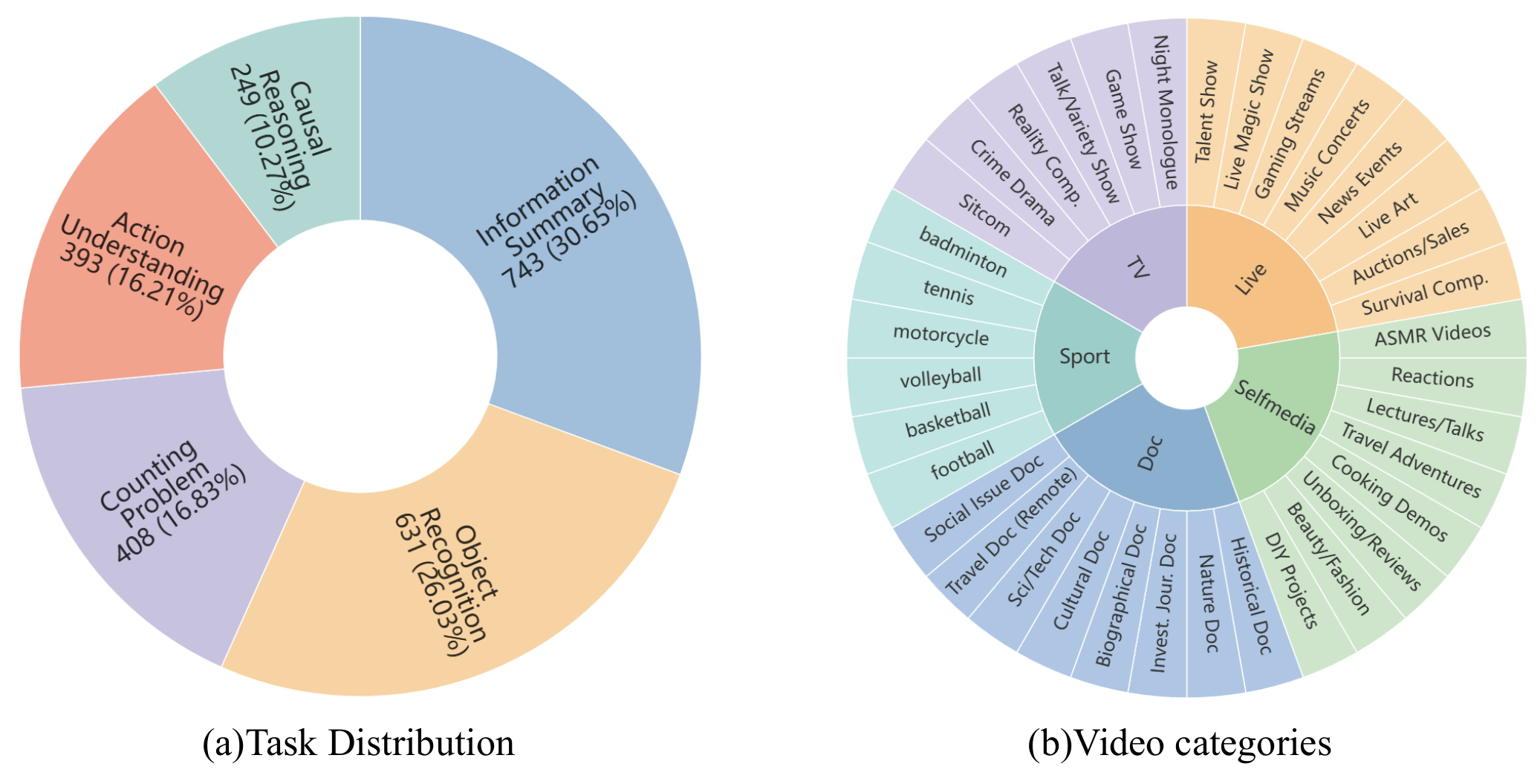}
    \caption{(a) Task distribution in LongVideoBench. LongVideoBench consists of a total of 5 tasks, ensuring comprehensive evaluation of the model's capabilities. (b) Video Categories. LongVideoBench includes videos spanning 5 major categories and 37 subcategories, ensuring diverse topical coverage. }
    \label{fig:8i}
\end{figure}

\newpage

\tableofcontents

\newpage

\section{Introduction}

Recent advances in Multimodal Large Language Models (MLLMs) have significantly enhanced their ability to integrate and interpret complex inputs such as text, images, and videos~\citep{liu2023visual, chen2024internvl, guo2024mammoth,zhang2024flash,zhu2025internvl3,team2025kimi}. Consequently, a variety of benchmarks have been developed to gauge their video understanding capabilities across different scopes and tasks~\citep{2024arXiv240509711W,2023arXiv230510683W,fu2024videommefirstevercomprehensiveevaluation,li2024videovistaversatilebenchmarkvideo,ma2025iv}.

However, current long-video benchmarks are ill-equipped to assess the multi-timescale capabilities of MLLMs—specifically, their distinct abilities across varying temporal granularities. By typically using isolated short segments~\citep{li2024videovista} or evaluating different temporal scales across entirely different videos~\citep{zhou2024mlvu}, these benchmarks inherently conflate temporal granularity with content variability. This makes it exceedingly difficult to disentangle a MLLM's true performance at each specific timescale from content-driven adaptations. Thus, a rigorous, fine-grained methodology is critically needed to evaluate how MLLMs apply these distinct temporal capabilities to understand the hierarchical temporal structures within long videos.

To bridge this gap, we introduce ScaleLong, a benchmark tailored for the fine-grained evaluation of MLLMs' multi-timescale capabilities in long videos. Its core feature is embedding questions targeting four hierarchical temporal scales (Clip, Shot, Event, Story) all within the same video content. This `within-content' design enables direct comparison of an MLLM's performance across these distinct temporal granularities on identical narratives, thereby isolating its abilities at each specific scale. ScaleLong includes 269 videos (averaging 86 minutes), each annotated with 4–8 questions, ensuring at least one question per time scale. As illustrated in Fig.~\ref{fig:8i}, the benchmark spans 5 major categories and 36 subcategories, enabling a comprehensive evaluation of MLLMs' understanding of long videos across diverse timescales.

Leveraging ScaleLong, extensive evaluations of 23 MLLMs—encompassing 19 open-source and 4 proprietary models—consistently reveal a U-shaped performance curve across the defined temporal scales. These models generally perform better on questions at the shortest (Clip) and longest (Story) temporal scales, while performance noticeably drops at intermediate levels (Shot and Event). This pattern suggests that current MLLMs often excel at processing either highly localized visual details or overarching narrative structures, yet face challenges with intermediate temporal contexts. Furthermore, targeted experiments conducted on ScaleLong indicate that an increased allocation of visual tokens systematically enhances MLLMs' performance across all evaluated timescales, providing valuable insights for future advancements in model development.

In summary, our work makes three primary contributions:

\begin{itemize}
\item \textbf{ScaleLong:} We introduce ScaleLong, specifically designed to assess the multi-timescale capabilities of MLLMs in long videos. By embedding questions at four hierarchical temporal scales (Clip, Shot, Event, and Story) within the same video content, it enables robust evaluation of MLLMs at each distinct scale. ScaleLong includes 269 diverse long videos (averaging 86 minutes), with 4-8 questions per video (at last one per scale), across 5 major categories and 36 subcategories.

\item \textbf{Comprehensive MLLM Evaluation and Insights}: Our extensive evaluation of 23 MLLMs on ScaleLong reveals a consistent U-shaped performance trend. MLLMs generally exhibit stronger comprehension at the shortest (Clip) and longest (Story) temporal scales, while their performance discernibly dips at intermediate scales (Shot and Event). This finding offers critical insights into how MLLMs process information at distinct temporal granularities in long videos.

\item \textbf{Insights for MLLM Development}: Evaluations on ScaleLong provide key insights for MLLM enhancement. For instance, increasing visual token allocation consistently enhances performance across all evaluated timescales. Furthermore, analysis of model error patterns reveals persistent weaknesses, offering guidance for future model improvements in long-video understanding.

\end{itemize}

\section{ScaleLong}

\subsection{Overview}
\begin{figure*}[t]
    \centering
    \includegraphics[width=0.90\textwidth]{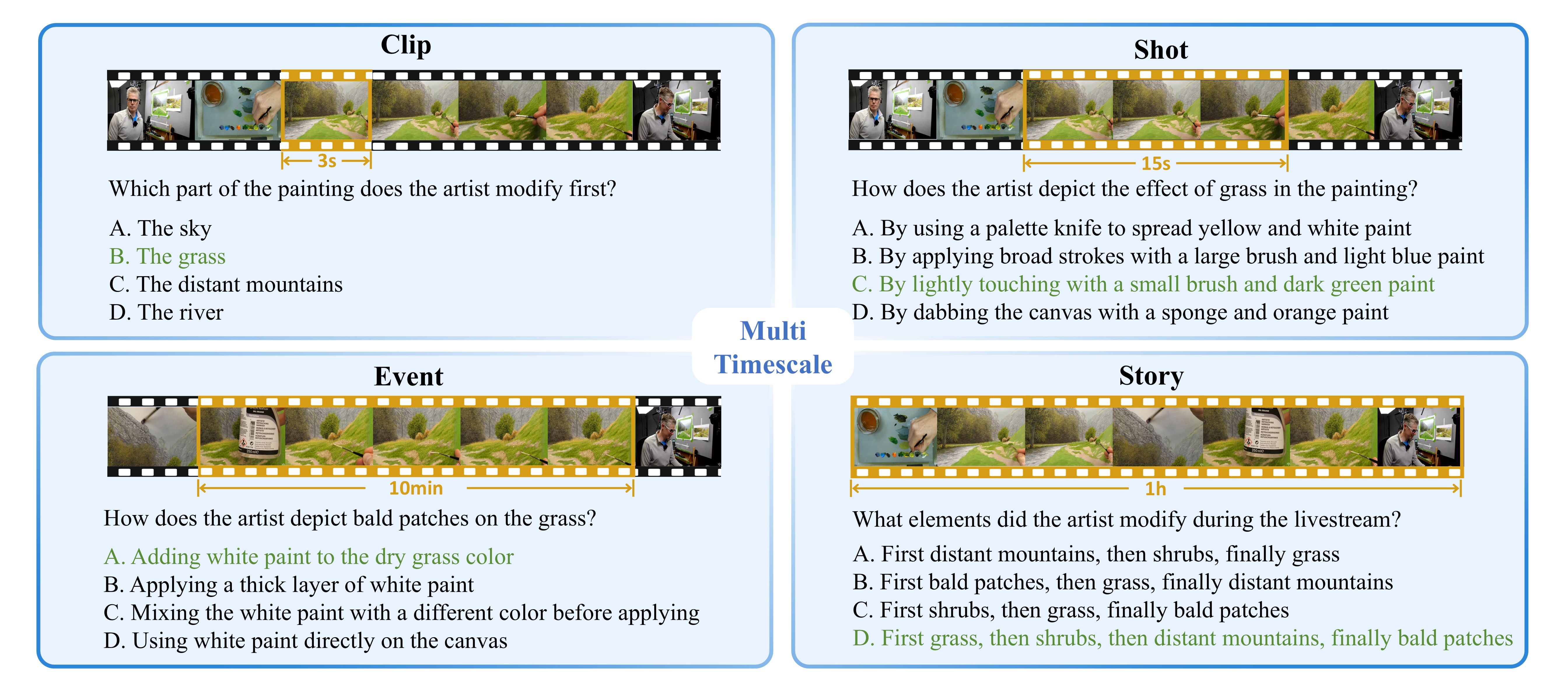}
    \caption{Representative samples from ScaleLong. Each sample in ScaleLong comprises a video paired with carefully designed questions, structured across four hierarchical temporal scales. The correct answers are indicated in green.}
    \label{fig:tasks}
\end{figure*}
ScaleLong is specifically engineered for the fine-grained assessment of the multi-timescale capabilities of MLLMs in long videos. The benchmark comprises 269 diverse videos, each averaging 86 minutes and annotated with 4-8 questions. As illustrated in Fig. \ref{fig:tasks}, these questions address four hierarchical temporal scales: Clip (seconds), Shot (tens of seconds), Event (minutes), and Story (hours). Key features of ScaleLong include:

\noindent \textbf{Multi Timescale Queries}: Unlike existing benchmarks, ScaleLong structures queries at four meticulously defined temporal scales—Clip, Shot, Event, and Story—all within each individual video. This `within-content' embedding of questions targeting multiple, distinct temporal scales is crucial: it effectively decouples the assessment of temporal understanding of specific video content. Such a design enables precise evaluation of how MLLMs handle different temporal granularities while keeping the narrative context consistent.

\noindent \textbf{Diverse Video Content and Task Design}: For comprehensive MLLM evaluation, ScaleLong offers extensive content diversity, featuring 5 main video categories (e.g., Sports, Documentaries) spanning 36 subcategories. It also incorporates 5 distinct task types (e.g., Causal Reasoning, Action Understanding) designed to probe deeper comprehension. This structured variety ensures representative assessment across diverse, real-world long-video scenarios.

\subsection{Multi-Timescale Hierarchies}

The Multi-Timescale Hierarchies within ScaleLong are established by categorizing questions into four distinct temporal levels. This classification is based on the video duration essential for answering each question and how relevant information is distributed across the frames. These levels are detailed as follows:

\noindent \textbf{Clip}: Questions solvable by analyzing a few consecutive frames, spanning only a few seconds (e.g., up to 3seconds), typically involving recognizing instantaneous actions, immediate visual details, or straightforward objects.

\noindent \textbf{Shot}: Questions requiring the analysis of information from multiple frames within a single continuous shot, typically ranging from 4 to 15 seconds. These questions require interpreting short-term dynamics, simple actions, character interactions, or semantic coherence within this timeframe.

\noindent \textbf{Event}: Questions concerning significant events that span multiple consecutive shots, with durations from 16 seconds up to 10 minutes. These questions require integrating information across scenes, interpreting event sequences, identifying contextually relevant frames, and understanding more complex narrative developments or causal links.

\noindent \textbf{Story}: Questions addressing content from the entire video or substantial portions thereof, typically exceeding 10 minutes. These require holistic comprehension of the overall narrative, including overall narrative logic, causal relationships, character development, thematic analysis, or long-term dependencies across the video.

\subsection{Task Types}

ScaleLong incorporates five distinct task types, each designed to rigorously evaluate different facets of an MLLM's comprehension abilities:

\noindent \textbf{Causal Reasoning (CR)}: Questions requiring inference about causal relationships within the video content. These tasks assess the model's ability to deduce cause-and-effect dynamics and logical connections.

\noindent \textbf{Object Recognition (OR)}: Questions involving the identification and distinction of specific objects, scenes, or their attributes within the video. These tasks evaluate visual perception and fine-grained recognition capabilities.

\noindent \textbf{Action Understanding (AU)}: Questions focused on interpreting the actions, movements, or behaviors of entities (e.g., characters, objects) in the video. These tasks assess the capacity to comprehend dynamic interactions and temporal movements.

\noindent \textbf{Information Summary (IS)}: Questions that require summarizing or generalizing main content, key points, or details from the video. These tasks evaluate the ability to synthesize information and extract essential concepts.

\noindent \textbf{Counting Problems (CP)}: Questions pertaining to quantitative aspects, such as enumerating objects or events, or discerning temporal order. These tasks assess the capacity for accurate numeric and sequential reasoning.

\subsection{Annotation Methodology and Quality Control}

\subsubsection{Annotation Methodology}

ScaleLong ensures high-quality annotations through a multi-phase process involving curated video selection, structured question design, and multi-round quality control. This process emphasizes content-based understanding, requiring questions to target video-specific information, answers to be thoroughly video-grounded, and dependencies on absolute time cues or external knowledge to be eliminated.

\noindent \textbf{Video Curation and Collection}: Our video acquisition process begins by defining 5 principal categories (further detailed into 36 subcategories) to cover diverse real-world scenarios. YouTube videos, typically around one hour in length, are manually sourced for these categories. Each selected video undergoes inspection for high visual clarity, substantial information density, and appropriate duration, resulting in a final corpus of 269 videos.

\noindent \textbf{Question, Answer, and Distractor Generation}: For each video, annotators first conduct a full viewing. They then design 8 questions (two for each of the four defined temporal hierarchy levels), ensuring a balance in task types. Correct answers are derived from careful analysis of multimodal information within the video. Each question is accompanied by one correct answer and three plausible distractors, which are constructed based on ten predefined types to offer varied challenges and facilitate error analysis.

\subsubsection{Rigorous Quality Control}

Our quality control protocol involves two principal rounds with distinct objectives:

\noindent \textbf{First-Round Quality Control}: This round focuses on the foundational correctness, clarity, and consistency of all annotations. Question stems are verified for precision. Critically, absolute time localizations are replaced with descriptive cues to compel content-based reasoning rather than timestamp reliance. Answer options are thoroughly checked: correct answers must be unambiguously video-grounded, and distractors plausible yet definitively incorrect.  Annotations also undergo checks to prevent an undue concentration of questions within limited segments of the video's timeline, and to validate all categorizations (e.g., temporal levels, task types, distractor types).

\noindent \textbf{Second-Round Quality Control}: This round focuses on eliminating confounding factors and ensuring questions exclusively assess understanding derived from the video content. Questions solvable through common world knowledge or reliant on external prior information, rather than video-specific details, are systematically revised or removed to nullify such external dependencies. Finally, to uphold dataset integrity, any questions exhibiting persistent ambiguities (e.g., unclear grounding, indistinct features, or problematic categorization) are rigorously discarded.

\section{ Comparison with other video benchmarks}
% \captionsetup{justification=raggedright}  % 设置标题左对齐
\begin{table*}[t]
\centering
\caption{Comparison with other benchmarks, where the abbreviations are defined as follows: \textbf{Anno.} (Annotation Method), \textbf{A} (Automatic Annotation), \textbf{M} (Manual Annotation), \textbf{\#Genres} (Number of Video Genres).  MTS is the abbreviation for Multi-Timescale, and IV-MTS is the abbreviation for Intra-Video Multi-Timescale.}

\label{tab:bench_comparison}
\resizebox{\textwidth}{!}{%
\begin{tabular}{ccccccccc}
\hline
\toprule
\textbf{Benchmark} & \textbf{\#Videos} & \textbf{Duration. (s)} & \textbf{\#Tasks} & \textbf{\#QA Pairs} & \textbf{Anno.} & \textbf{\#Genres} & \textbf{MTS} & \textbf{IV-MTS} \\ 
\midrule
MSVD-QA & 1,970 & 10 & - & 13,157 & A & -  & \textcolor{red}{\ding{55}} & \textcolor{red}{\ding{55}} \\
MSRVTT-QA & 2,900 & 15 & - & 72,821 & A & - & \textcolor{red}{\ding{55}} & \textcolor{red}{\ding{55}} \\
ActivityNet-QA & 5,800 & 111 & 4 & 800 & M & 8  & \textcolor{red}{\ding{55}} & \textcolor{red}{\ding{55}}  \\
NExTQA & 1,000 & 44 & 4 & 8,564 & M & - & \textcolor{red}{\ding{55}} & \textcolor{red}{\ding{55}}  \\
MVBench & 3,641 & 16 & 20 & 4,000 & A & - & \textcolor{red}{\ding{55}} & \textcolor{red}{\ding{55}}  \\
CinePile & 9,396 & 160 & 5 & 303,828 & M \& A & 1 & \textcolor{red}{\ding{55}} & \textcolor{red}{\ding{55}}  \\
EgoSchema & 5,063 & 180 & - & 5,063 & M \& A & - & \textcolor{red}{\ding{55}} & \textcolor{red}{\ding{55}}  \\ 
\midrule
LVBench & 103 & 4,101 & 6 & 1,549 & M & 21 & \textcolor{red}{\ding{55}} & \textcolor{red}{\ding{55}} \\
LONGVIDEOBENCH & 3,763 & 473 & 17 & 6,678 & M  & 10 & \textcolor{red}{\ding{55}} & \textcolor{red}{\ding{55}} \\
HourVideo & 500 & 2,742 & 4 & 12,976 & M \& A & - & \textcolor{red}{\ding{55}} & \textcolor{red}{\ding{55}} \\
ALLVB & 1,376 & 7,200 & 9 & 252,000 & A & 16 & \textcolor{red}{\ding{55}} & \textcolor{red}{\ding{55}} \\
Video-MME & 900 & 1,024 & 12 & 2,700 & M & 30 & \textcolor{green}{\ding{51}} & \textcolor{red}{\ding{55}}  \\
MoVQA & 100 & 992 & 6 & 21,953 & M & 1 & \textcolor{green}{\ding{51}} & \textcolor{red}{\ding{55}}  \\
MLVU & 1,730 & 930 & 9 & 3,102 & M & 31 & \textcolor{green}{\ding{51}} & \textcolor{red}{\ding{55}} \\
% HLV-1K & 1,009 & 3,300 & 15 & 14,847 & M \& A & 9 & \textcolor{green}{\ding{51}} & \textcolor{green}{\ding{51}} \\
\midrule
\textbf{ScaleLong} & 269 & 5,160 & 5 & 1747 & M & 36 & \textcolor{green}{\ding{51}} & \textcolor{green}{\ding{51}} \\
\bottomrule
\end{tabular}}

\end{table*}

Existing video understanding benchmarks, as detailed in Table \ref{tab:bench_comparison}, are broadly categorized into short-video and long-video formats. Short-video benchmarks like NExTQA~\citep{xiao2021next} and MVBench~\citep{li2024mvbench} utilize sub-minute clips, which restrict their capacity for evaluating long-range temporal understanding. Recent long-video benchmarks—including CinePile~\citep{rawal2024cinepile}, EgoSchema~\citep{mangalam2023egoschema}, MoVQA~\citep{zhang2023movqa}, MLVU~\citep{zhou2024mlvu}, Video-MME~\citep{fu2024video}, LongVideoBench~\citep{wu2024longvideobenchbenchmarklongcontextinterleaved}, and ALLVB~\citep{tan2025allvb}—feature extended durations. However, they generally do not decouple the targeted temporal scales of questions from specific video content. This inherent coupling hinders a precise assessment of how MLLMs handle varying temporal granularities and, consequently, their distinct multi-timescale capabilities.

Unlike most existing long-video benchmarks which, as noted, typically conflate temporal scale assessment with disparate video content, ScaleLong is distinguished by a key design attribute: its Intra-Video Multi-Timescale nature. This principle dictates that within every long video, questions are specifically designed to target multiple, distinct temporal scales. This inherent characteristic—materialized through questions probing four hierarchical levels (\emph{Clip} (seconds), \emph{Shot} (tens of seconds), \emph{Event} (minutes), and \emph{Story} (hours)) all within the same video narrative—is fundamental to the engineering of ScaleLong for the precise assessment of multi-timescale capabilities of MLLMs in Long Videos. Such a design feature directly facilitates a disentangled evaluation; the distinct capabilities of an MLLM at various temporal granularities are thereby measured against the same video content, allowing for a clear separation of temporal understanding performance from content-specific reactions.

\section{Experiments}\label{sec:experiments}
% In this section, we conduct a comprehensive evaluation of several representative MLLMs using ScaleLong. We begin by introducing the selected models and experimental setup, followed by a quantitative comparison of overall performance between open-source and commercial models. We then analyze three key factors that influence model performance across different temporal reasoning levels: the impact of frame count, the effect of resolution, and the trade-off between frame count and resolution. Finally, we analyze error rates across different distractor types, revealing key vulnerabilities and failure patterns in complex reasoning scenarios.

In this section, we evaluate representative MLLMs on ScaleLong—first outlining the setting of experiments,  then analyzing the results of the 23 MLLMs on ScaleLong. We assess how the visual tokens shape long video understanding, and finally examine error rates by distractor type to identify key failure modes.

\subsection{Settings}
% We evaluate 4 commercial models: GPT-4o~\citep{gpt4o}, Gemini-2.0-flash~\citep{team2024gemini}, Gemini-2.5-pro~\citep{gemini25pro2025}, and Doubao-1.5-vison-pro~\citep{doubao1.5pro}. For open-source evaluation, we select 18 representative MLLMs, including the Qwen2.5-VL~\citep{Qwen2.5-VL}, InternVL2.5~\citep{chen2024expanding}, and LLaVA-OneVision~\citep{llava}, covering model scale from 7B to 78B. ScaleLong, which consists of single-answer multiple-choice questions with one correct option and three distractors covering diverse reasoning challenges. We adopt accuracy as the primary evaluation metric, defined as the percentage of questions for which the model’s prediction matches the ground-truth answer.

We evaluate a total of 23 MLLMs, comprising 4 leading commercial models—Gemini-2.5-pro~\citep{gemini25pro2025}, Gemini-2.0-flash~\citep{team2024gemini}, GPT-4o~\citep{gpt4o} and Doubao-1.5-vision-pro~\citep{doubao1.5pro}—and 19 open-source models spanning from 7 billion to 78 billion parameters, including representative models such as Qwen2.5-VL~\citep{Qwen2.5-VL}, InternVL2.5~\citep{chen2024expanding} and LLaVA-OneVision~\citep{llava}.

\subsection{Main Results}

\begin{table*}[t]
\centering
\caption{The performance of proprietary and open-source MLLMs on ScaleLong across granularities and task types. For each timescale and task, the best performance is indicated in bold, and the second-best performance is indicated with underlining.}
  \label{tab:llm_comparison}
\resizebox{\textwidth}{!}{%
  \begin{tabular}{lc c |c c c c| c c c c c|c}
    \toprule
    \multirow{2}{*}{\textbf{Models}} & \multirow{2}{*}{\textbf{Date}} & \multirow{2}{*}{\textbf{Input}} & \multicolumn{4}{c|}{\textbf{Granularities}} & \multicolumn{5}{c|}{\textbf{Task Types}} & \multirow{2}{*}{\textbf{Overall}}\\
    \cmidrule(lr){4-7} \cmidrule(lr){8-12}
 & &&  \textbf{Clip} & \textbf{Shot} & \textbf{Event} & \textbf{Story} & \textbf{CR} & \textbf{OR} & \textbf{AU} & \textbf{II} & \textbf{CP} \\
     \midrule
\multicolumn{13}{l}{\textit{Proprietary Models}}\\
\midrule
Gemini 2.0 Flash & 2025-02 & 256 frm & 65.7 & 52.4 & 48.4 & 53.4 & 53.5 & 64.8 & 54.6 & 55.9 & 41.9 & 55.0 \\
GPT-4o & 2024-05 & 64 frm & 61.8 & 50.7 & 51.0 & 58.0 & \underline{58.3} & 62.6 & 57.4 & 60.1 & 36.0 & 55.4 \\
Doubao 1.5-VL Pro & 2025-01 & 256 frm & \underline{66.4} & 52.8 & \underline{55.2} & 60.2 & 57.1 & \underline{67.0} & 55.1 & \underline{64.5} & 43.3 & \underline{58.7} \\
Gemini 2.5 Pro & 2025-03 & 256 frm & \textbf{71.5} & \textbf{62.8} & \textbf{68.0} & \textbf{69.0} & \textbf{66.0} & \textbf{72.5} & \textbf{65.8} & \textbf{74.6} & \textbf{51.2} & \textbf{67.9} \\

\midrule
\multicolumn{13}{l}{\textit{Open source Models}}\\
\midrule
% \midrule
% \multicolumn{12}{l}{\textit{Open Source MLLMs($<$ 10B)}} \\
% \midrule
LLaVA-Mini & 2025-01 & 256 frm & 29.7 & 25.3 & 28.8 & 25.2 & 27.6 & 29.8 & 29.4 & 27.2 & 22.1 & 27.3 \\
LongVILA   & 2024-08 & 32 frm & 29.1 & 28.3 & 23.8 & 28.6 & 28.0 & 29.2 & 30.0 & 26.8 & 23.9 & 27.5 \\
LongVU     & 2024-10 & 32 frm & 40.9 & 37.2 & 33.5 & 35.6 & 43.9 & 44.1 & 37.5 & 38.1 & 21.7 & 36.8 \\
Phi-3.5    & 2024-04 & 64 frm & 44.8 & 35.8 & 34.3 & 43.0 & 43.3 & 47.9 & 33.9 & 40.1 & 30.2 & 39.5 \\
LongVA     & 2024-06 & 256 frm & 50.3 & 40.2 & 38.8 & 43.8 & 49.3 & 53.1 & 38.2 & 45.4 & 28.8 & 43.3 \\
% MiniCPM-O  & 2024-08 & 32 frm & 47.3 & 40.0 & 39.5 & 44.8 & 45.2 & 49.6 & 40.0 & 48.6 & 27.2 & 42.9 \\
Flash-VStream & 2024-06 & 256 frm & 46.9 & 42.7 & 39.8 & 47.0 & 48.7 & 48.9 & 39.0 & 34.1 & 48.6 & 44.1 \\
Phi-4      & 2025-03 & 128 frm & 50.0 & 42.9 & 42.1 & 45.5 & 50.3 & 53.9 & 44.0 & 46.2 & 30.8 & 45.2 \\
LLaVA-OV-7B(SI) & 2024-08 & 128 frm & 50.8 & 39.5 & 44.4 & 47.3 & 41.4 & 52.2 & 44.1 & 49.3 & 34.3 & 45.5 \\
MiniCPM-V       & 2024-08 & 64 frm & 51.0 & 42.9 & 43.1 & 47.8 & 49.0 & 55.2 & 42.9 & 48.5 & 32.5 & 46.2 \\
Qwen2-VL-7B     & 2024-09 & 8 frm  & 51.6 & 45.9 & 46.5 & 48.3 & 51.6 & 51.9 & 53.1 & 50.6 & 33.9 & 48.1 \\
LLaVA-Video-7B     & 2024-10 & 32 frm & 57.9 & 46.8 & 50.2 & 48.0 & 47.1 & 55.4 & 52.3 & 54.4 & 39.9 & 50.8\\
InternVL2-5-8B  & 2024-12 & 128 frm & 60.2 & 42.5 & 48.5 & 52.3 & 52.0 & 61.5 & 47.4 & 50.8 & 39.3 & 50.9 \\
Qwen2.5-VL-7B   & 2025-01 & 256 frm & 52.8 & 50.0 & 49.5 & 52.7 & 50.2 & 54.5 & 53.3 & 55.3 & 39.9 & 51.2 \\
% \midrule
% \multicolumn{12}{l}{\textit{Open Source MLLMs($<$ 10B)}} \\
% \midrule
Aria            & 2024-10 & 256 frm & 57.2 & 46.5 & 48.7 & 53.4 & 49.3 & 60.5 & 48.5 & 52.4 & 41.6 & 51.5 \\
LLaVA-OV-72B & 2024-08 & 64 frm & 56.1 & 49.5 & 51.5 & 55.2 & 53.6 & 59.3 & 53.5 & 52.3 & 45.5 & 53.1\\
InternVL2-5-26B & 2024-12 & 128 frm & 60.2 & 50.1 & 48.5 & 56.8 & 57.9 & 61.7 & 53.6 & 52.9 & 43.6 & 53.9 \\
LLaVA-Video-72B & 2024-10 & 128 frm & 60.0 & 50.7 & 53.2 & 51.8 & 54.6 & 62.7 & 55.5 & 55.2 & 39.0 & 53.9 \\
InternVL2-5-38B & 2024-12 & 256 frm & 61.8 & 53.5 & 54.1 & 55.5 & 53.9 & 65.4 & \underline{58.8} & 58.1 & 40.7 & 56.3 \\
InternVL2-5-78B & 2024-12 & 128 frm & 65.2 & \underline{54.3} & 53.4 & \underline{61.5} & 57.2 & 65.4 & 54.4 & 63.9 & \underline{46.2} & 58.6 \\

\bottomrule
\end{tabular}%
}
\end{table*}
\vspace{1em}

Table~\ref{tab:llm_comparison} presents the performance of all evaluated models across four timescale questions as well as five task types defined in ScaleLong. In this subsection, for all experiments, we fix the resolution at 240p and use the highest frame count we have tested. We draw the following key observations:

\noindent \textbf{Model performance varies significantly across timescales:} In long-video understanding, we evaluate models across four timescales—Clip (shortest span), Shot and Event (mid-range spans), and Story (longest span). We observe a pronounced U-shaped trend: accuracy peaks at the two extremes (Clip and Story) but dips markedly at the intermediate timescales (Shot and Event). This indicates that, while current MLLMs excel at capturing brief visual cues and overarching narrative structures, they struggle with maintaining temporal coherence over moderate-length segments. For example, Gemini 2.5 Pro achieves 71.5 \% accuracy on Clip and 69.0\% on Story, yet drops to 62.8 \% on Shot and 68.0 \% on Event. Crucially, this U-shaped pattern holds consistently across all open-source and closed-source models.

\noindent \textbf{Performance differences across models are notable:} Closed-source models consistently outperform open-source ones across all timescales. For example, the leading closed-source model, Gemini 2.5 Pro, surpasses the best open-source counterpart (InternVL2.5-78B) by at least 6.3 percentage points on the Clip timescale and by 14.6 points on the Event timescale. Furthermore, within the InternVL2.5 series, scaling from 8 B to 78 B yields steady accuracy gains: from 60.2 \% to 65.2 \% on Clip, 52.3 \% to 61.5 \% on Story, 42.5 \% to 54.3 \% on Shot, and 48.5 \% to 53.4 \% on Event.

\noindent \textbf{MLLMs exhibit substantial performance disparities across task types:} For the vast majority of models, Object Recognition tasks achieve the highest accuracy, whereas Counting Problems tasks incur the lowest. For example, Doubao 1.5-VL Pro shows a 23.7 percentage-point gap between OR and CP, while GPT-4o exhibits a 26.6-point difference. This consistent gap underscores that, in long-video understanding, MLLMs' ability to perform counting remains a critical area for improvement.

\begin{figure*}[htbp]
    \centering
    \begin{subfigure}[b]{0.33\textwidth}
        \centering
        \includegraphics[width=\textwidth]{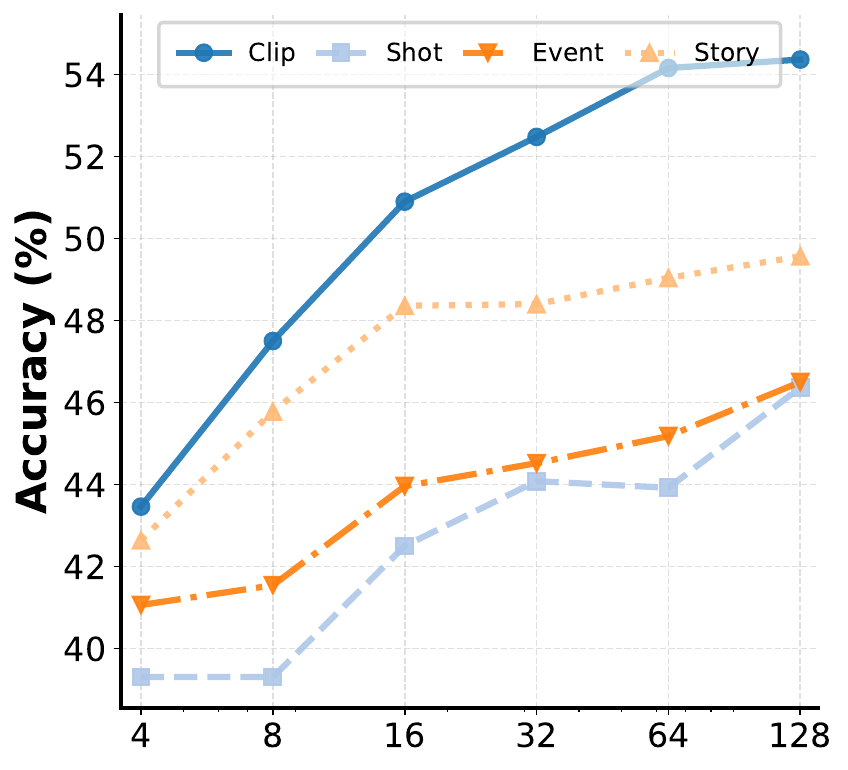}
        \captionsetup{justification=centering} % 让 caption 居中
        \caption*{(a)} % 避免自动编号，仅显示 (a)
        \label{fig:frames-high}
    \end{subfigure}
    \hfill
    \begin{subfigure}[b]{0.31\textwidth}
        \centering
        \includegraphics[width=\textwidth]{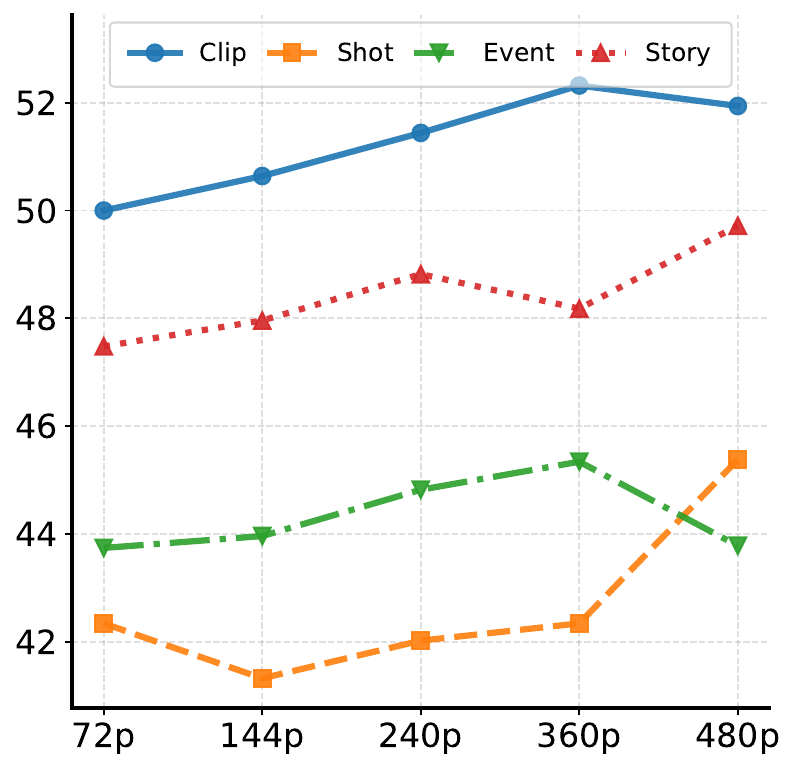}
        \captionsetup{justification=centering} % 让 caption 居中
        \caption*{(b)} % 避免自动编号，仅显示 (b)
        \label{fig:resolutions2}
    \end{subfigure}
    \hfill
    \begin{subfigure}[b]{0.32\textwidth}
        \centering
        \includegraphics[width=\textwidth]{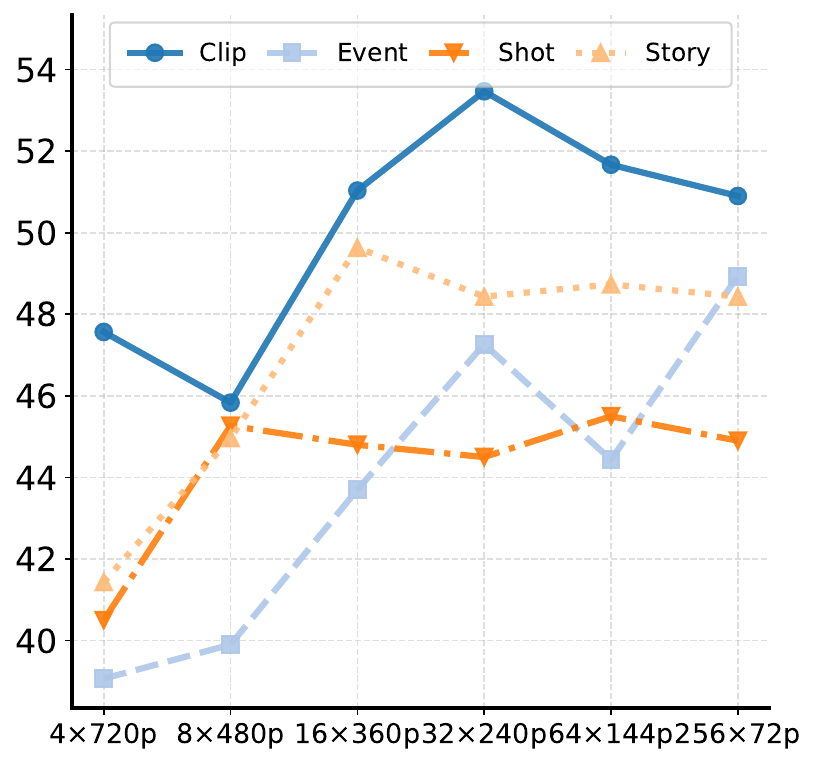}
        \captionsetup{justification=centering} % 让 caption 居中
        \caption*{(c)} % 避免自动编号，仅显示 (c)
        \label{fig:trade-off1-high}
    \end{subfigure}

    \caption{Comparison of model performance under: (a) varying frame counts, (b) varying video resolutions, and (c) different frame–resolution combinations.}
    \label{fig:comparison1}
\end{figure*}

\subsection{Ablation Study}

\noindent \textit{To investigate how total visual‐token count and its allocation between frame number and resolution affect MLLM performance in multi‐timescale long‐video understanding}, we conduct two ablation studies:

\noindent\textbf{Scaling Effect.} How does performance change as we increase the total number of visual tokens—either by sampling more frames or by raising resolution?

\noindent\textbf{Token Allocation.} When the total visual‐token budget is held constant, does distributing tokens across more frames or into higher resolution yield greater gains?

\subsubsection{Isolated Scaling of Frame number and Resolution}

In this subsection, we evaluate how allocating extra visual tokens—temporally by sampling more frames or spatially by increasing resolution—affects model performance.  

% We conduct experiments on six top-performing models—Aria, MiniCPM-O, InternVL-8B/26B, and Qwen2.5-VL-7B/72B—under two settings. 

\noindent \textbf{Under a fixed resolution, increasing the number of input frames consistently improves multi-timescale long-video understanding, with the greatest gains on Clip-level tasks.} In this section, we evaluate several representative MLLMs using input frame counts of 4, 8, 16, 32, 64, and 128 with 240P resolution. The corresponding results are presented in Fig.~\ref{fig:comparison1}(a). Overall, the accuracy of all models tends to improve as the number of input frames increases. However, the degree of improvement varies significantly across different levels of temporal granularity. The Clip level exhibits the most substantial gain, with accuracy increasing from approximately 43.5\% at 4 frames to 54.5\% at 128 frames. This suggests that short-span tasks are highly sensitive to temporal sampling density, likely because they depend on capturing fine-grained visual changes or brief actions. In contrast, the accuracy improvements at the Shot and Event levels are more moderate. Notably, accuracy for the Event level peaks at 64 frames and slightly declines at 128 frames, indicating potential redundancy or oversaturation when too many frames are included. For the Story level, performance gains are relatively limited, increasing from around 42.5\% to 49.5\%. This implies that for long-range reasoning tasks, a small number of well-chosen frames may already provide sufficient contextual information, and further increasing the frame count yields diminishing returns.

In terms of resolution, we evaluate model performance using 32 frames across five video resolutions (72p, 144p, 240p, 360p, and 480p). As shown in Fig. \ref{fig:comparison1}(b), we observe that:

\noindent \textbf{Under a fixed frame count, raising resolution generally improves performance across Clip, Shot, Event, and Story tasks, but sometimes yields diminishing or even negative returns.} For most timescales, model accuracy climbs as resolution increases—for example, at the Clip level, moving from 72 p to 360 p delivers roughly a 2 \% absolute gain in accuracy, and Story-level tasks show a similar uplift. However, when viewed in an absolute sense, boosting resolution proves less effective than increasing frame count: earlier experiments demonstrate that, at Clip granularity, expanding input frames from 4 to 128 yields about a 9 \% jump in accuracy—substantially more than 2 \% gain from resolution alone.

That said, resolution gains are not strictly monotonic. At the Clip level, accuracy actually dips slightly when stepping from 360 p to 480 p, suggesting that excessive spatial detail can introduce noise or redundant information that marginally hinders short-span reasoning.

\subsubsection{Token Allocation: Frames vs. Resolution} 

To disentangle temporal and spatial contributions under a fixed visual‐token budget, we evaluate Qwen2.5-VL-7B/32B/72B on six frame–resolution combinations: 4×720p, 8×480p, 16×360p, 32×240p, 64×144p, and 256×72p. We report the average accuracy of three models.

\noindent \textbf{Appropriate allocation of visual tokens between temporal and spatial dimensions is vital for multi‐timescale long‐video understanding.} As shown in Fig.~\ref{fig:comparison1}(c), Clip-level accuracy peaks at 32×240p (53.6\%), highlighting the value of temporal density for short-span tasks. Story-level accuracy gradually improves and reaches its peak (49.8\%) at the 16×360p setting. Beyond this point, performance slightly declines and then stabilizes, indicating diminishing returns from further increasing frame count or decreasing resolution. Event-level accuracy also benefits from additional frames, peaking at 48.8\%, though low resolution (e.g., 64×144p) can cause instability. Notably, Shot-level accuracy starts low (40.2\%) and improves sharply at 8×480p (45.5\%), then plateaus, indicating that a moderate balance of temporal and spatial input is most effective.

% models perform best when frame count and resolution are balanced. This suggests that extremely low frame rates or resolutions significantly impair performance, and simply reallocating all visual tokens to the other dimension (e.g., favoring resolution over frame count) cannot fully compensate for the loss. 
\begin{figure*}[htbp]
    \centering
    \begin{subfigure}[b]{1\textwidth}
        \centering
        \includegraphics[width=\textwidth]{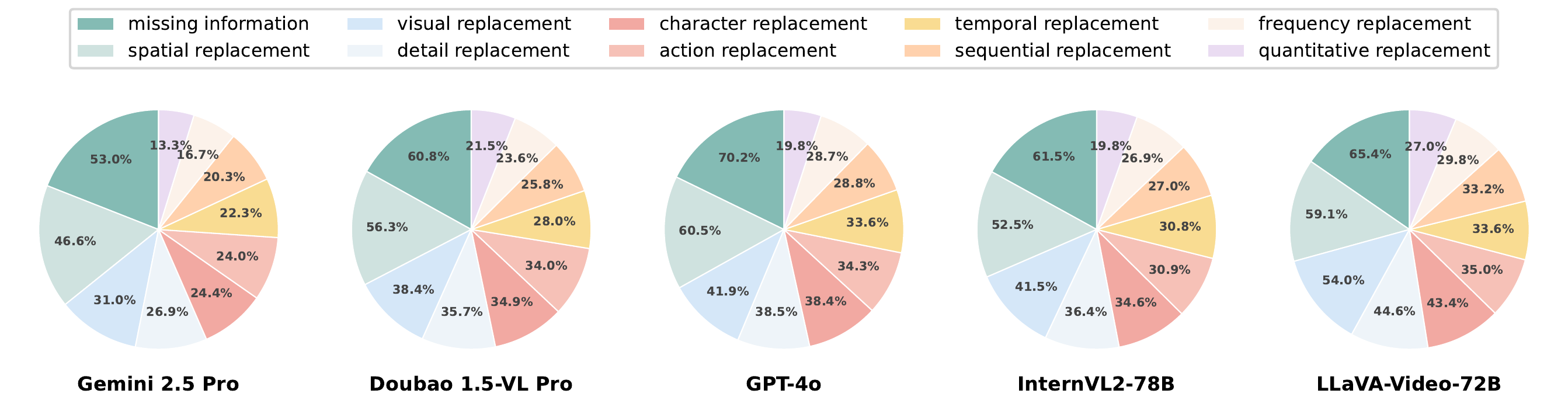}
        \captionsetup{justification=centering} % 让 caption 居中
        % \caption*{(c)} % 避免自动编号，仅显示 (b)
        \label{fig:wrong_design1}
    \end{subfigure}
    \caption{Distractor-specific error rate distribution across five MLLMs.}
    \label{fig:wrong_design}
\end{figure*}

\subsection{Error rates across different distractor types}

To analyze error patterns in long-video understanding, we evaluate several MLLMs on the ten distractor types in ScaleLong, as shown in Fig.~\ref{fig:wrong_design}. Although overall error rates are comparable across models, two categories—missing information and spatial replacement—stand out with the highest failure rates. For example, Gemini 2.5 Pro, our best-performing model, erroneously accepts missing-information distractors 53\% of the time and spatial-replacement distractors 46.6\% of the time. These findings indicate a pervasive insensitivity to the completeness of evidential support, as well as a notable deficiency in reasoning about spatial relationships within complex video sequences.

In contrast, models exhibit markedly stronger performance on frequency misdirection and quantitative misdirection. GPT-4o misclassifies these distractors only 19.8\% and 28.7\% of the time, respectively, while Gemini 2.5 Pro's error rates are even lower (13.3\% and 16.7\%). This suggests that, despite their struggles with semantic completeness and spatial inference, current MLLMs are adept at leveraging statistical and numerical cues. Together, these results highlight the need for future work to incorporate mechanisms—either through architectural enhancements or targeted training curricula—that explicitly verify evidential completeness and model multi-view spatial configurations in long-video contexts.

\section{Related Work}

\paragraph{Multimodal Large Language Models}

Multimodal LLMs (MLLMs) pair visual encoders with large language models to excel at image and short-video tasks (e.g., LLaVA-onevision~\citep{li2024llavaonevisioneasyvisualtask}, Otter~\citep{li2023ottermultimodalmodelincontext}, mPLUG-Owl~\citep{ye2023mplug}). For long-video understanding, recent MLLMs introduce specialized designs—Video-LLaMA~\citep{zhang2023video} (ViT + Q-Former), LLaMA-Vid~\citep{li2024llama} (efficient visual compression), mPLUG-owl3~\citep{ye2024mplug} (scalable multi-event modeling), LLaVA-Octopus~\citep{zhao2025llava} (audio integration)—and leverage expanded multimodal pretraining in InternVL2.5~\citep{chen2024expanding} and Qwen2.5-VL~\citep{Qwen2.5-VL}.

\paragraph{Video Understanding Benchmarks}

Video benchmarks have evolved from short‐clip tests (e.g., MVBench~\citep{li2024mvbench}, NExT-QA~\citep{xiao2021next}) through mid‐length tasks (CinePile~\citep{rawal2024cinepile}, EgoSchema~\citep{mangalam2023egoschema}, MoVQA~\citep{zhang2023movqa}, MLVU~\citep{zhou2024mlvu}, Video-MME~\citep{fu2024video}) to hour‐scale evaluations (LVBench~\citep{wang2024lvbench}, LONGVIDEOBENCH~\citep{wu2024longvideobenchbenchmarklongcontextinterleaved}, HourVideo~\citep{chandrasegaran2024hourvideo}, ALLVB~\citep{tan2025allvb}, HLV-1K~\citep{zou2025hlv}). However, current long‐video benchmarks are ill‐equipped to assess the multi‐timescale capabilities of multimodal LLMs—specifically, their distinct abilities across varying temporal granularities. Existing benchmarks rely on isolated clips or distribute scales across different videos, conflating temporal granularity with content variability and obscuring true model performance.
To address these limitations, ScaleLong embeds balanced question sets—complete with varied distractors—at clip, shot, event and story levels within the same hour-long videos drawn from diverse genres (documentaries, dramas, tutorials). This within-content, multi-scale design enables precise cross-granularity evaluation, revealing MLLMs’ accuracy trends and specific failure modes across the temporal hierarchy.
% 佳恒

\section{Conclusion}

We introduce ScaleLong, the first benchmark for fine-grained MLLM evaluation across hierarchical temporal scales (Clip to Story, spanning seconds to hours) using questions embedded within the same video. This 'within-content' design disentangles temporal scale effects from video semantics, enabling a more accurate assessment of intra-video multi-timescale understanding. Evaluations on 23 MLLMs reveal a U-shaped performance curve: models perform better at the shortest (Clip) and longest (Story) scales, with a dip at intermediate (Shot, Event) levels. Furthermore, visual token ablation studies indicate that under fixed budgets, a balanced allocation between frame count and resolution is optimal, as severe deficiency in one aspect significantly impairs performance. We hope ScaleLong will catalyze research to advance MLLM capabilities in nuanced, multi-scale long-video understanding.

\clearpage

\section{Contributions and Acknowledgments}

Multimodal Art Projection (M-A-P) is a non-profit open-source AI research community, run by donations.
The community members are working on research topics in a wide range of spectrum, including but not limited to the pre-training paradigm of foundation models, large-scale data collection and processing, and the derived applications on coding, reasoning, and music generation.

\textbf{Leading Authors}
\begin{multicols}{2}
    \begin{itemize}
        \item David Ma, M-A-P
        \item Huaqing Yuan
        \item Xingjian Wang
        \item Qianbo Zang, M-A-P
    \end{itemize}
\end{multicols}

\textbf{Contributors}
\begin{multicols}{2}
    \begin{itemize}
        \item Tianci Liu
        \item Xinyang He
        \item Zhenzhu Yang
        \item Yanbin Wei
        \item Jiawei Guo, M-A-P
        \item Jiahui Ni
        \item Zhenzhu Yang, M-A-P
        \item Meng Cao, MBZUAI
        \item Shanghaoran Quan, M-A-P
        \item Yizhi LI, M-A-P
        \item Wangchunshu Zhou, OPPO, M-A-P
        \item Jiaheng Liu, M-A-P, NJU
        \item Wenhao Huang, M-A-P
    \end{itemize}
\end{multicols}

\textbf{Corresponding Authors}
\begin{multicols}{2}
    \begin{itemize}
        \item Ge Zhang, M-A-P
        \item Shiwen Ni, SIAT-CAS
        \item Xiaojie Jin, M-A-P
    \end{itemize}
\end{multicols}

\newpage

\bibliography{main}
\newpage
\appendix

\newtcolorbox[auto counter, number within=section]{methodbox}[2][]{%
  colback=white, 
  colframe=teal!80!green!80!black,  
  width=\textwidth,
  arc=2mm, 
  boxrule=0.5mm, 
  title={\normalsize\faWrench\hspace{0.5em}#2}, 
  breakable,
  fonttitle=\bfseries\Large, 
  fontupper=\small, 
  % drop shadow south east, 
  % drop shadow={south east, teal!50!black, shadow xshift=1.5mm, shadow yshift=-1.5mm},
  % drop shadow={south east, teal!50!black},
  % drop shadow={south east, teal!50!black, opacity=0.8, shadow xshift=2mm, shadow yshift=-2mm},
  #1
}

\section{Annotation Tutorial}
\label{appendix: annotation tutorial}
\subsection{Question Type}
\begin{methodbox}{Question Type Details}
Representative examples of all 5 task categories in ScaleLong are shown in Figure~\ref{fig:tasks}.
\subsection*{Causal Reasoning}
\begin{itemize}
\item These questions aim to test the model's ability to infer causal relationships between events, actions, or phenomena in the video. The model needs to understand the content of frames over a certain time segment, and identify the internal logic of the “cause-effect” chain.
\end{itemize}
\subsection*{Object Recognition}
\begin{itemize}
\item These questions aim to assess the model's ability to identify specific objects, scenes, and their features (such as color, shape, and state) in the video. The model needs to locate them within the video scenes, achieve cross-frame tracking and consistent recognition.
\end{itemize}
\subsection*{Action Understanding}
\begin{itemize}
\item These questions aim to test the model's ability to identify character actions or object movements in the video. The model needs to understand the temporal combination of actions and their semantic goals.
\end{itemize}
\subsection*{Information Summary}
\begin{itemize}
\item These questions aim to test the model's ability to summarize or generalize the main content or details of the video. The model needs to use clues or context within the video to go beyond the understanding of individual frames or segments, grasp the core content of the video, and extract the plot summary.
\end{itemize}
\subsection*{Counting Problems}
\begin{itemize}
\item These questions aim to assess the model's ability to conduct quantitative analysis of the number of objects, frequency of events, and temporal relationships in the video. The model needs to accurately identify and distinguish various elements, involving counting across multiple dimensions such as objects, plot elements, or actions.
\end{itemize}
\end{methodbox}

\subsection{Data Annotation Steps}
\begin{methodbox}{Operations:}
\begin{enumerate}
\item \textbf{Attributes that need to be annotated:} 
\begin{itemize}
    \item \textbf{Video Key:} The video id in Youtube.
\end{itemize}
\begin{itemize}
    \item \textbf{Video Type:} Based on the content of the video, select one category from TV, sport, live, self-media, and documentary.
\end{itemize}
\begin{itemize}
    \item \textbf{Question Stems and Options:} Use clear and concise language to describe the question, ensuring that the question stem is explicit and specific. Each QA should include 4 options that are logically coherent and relevant to the question.
\end{itemize}
\begin{itemize}
    \item \textbf{Answer:} Provide a unique and correct answer.
\end{itemize}
\begin{itemize}
    \item \textbf{Question Type:} Select one of the 5 question types that align with the question stem, each question can have only one  question type.
\end{itemize}
\begin{itemize}
    \item \textbf{Time Reference:} Label the time segment in the video corresponding to the answer (the time segment format should be in string format "XX:XX-XX:XX").
\end{itemize}
\begin{itemize}
    \item \textbf{Hierarchy:} Label the level to which the question belongs (clip, shot, event, or story).
\end{itemize}
\item \textbf{Watch Video and Determine Hierarchy  Type:}

\begin{itemize}
\item After fully viewing the video content to be annotated, select the two most appropriate and valuable question types for each hierarchy. Then, pre-conceive the corresponding question content in preparation for designing the question stems and distractors.
\end{itemize}
\item \textbf{Design Question Stem and Answer:}
\begin{itemize}
\item \textbf{Question Design Requirements:}
\begin{itemize}
 \item \textbf{Clear Expression:} Ensure that the questions are concise and straightforward, avoiding complex or lengthy expression. 
\item \textbf{Explicit Description:} Describe the core elements in the video clearly and specifically (such as scenes, characters, objects, actions, or weather), ensuring questions are unambiguous and refer to a unique segment in the video. 
\item \textbf{Quantity Requirement:} For each video, design and annotate 2 questions for each hierarchy (clip, shot, event, story), need a total of 8 questions per video.
\item \textbf{Balanced Question Types:} Maintain a similar number of questions for each type (e.g., Causal Reasoning, Object Recognition, etc.).
\item \textbf{Target Visual Information:} Ensure that pure text-based LLMs cannot answer the questions correctly.
\end{itemize}
\item \textbf{Answer Design Requirements:}
\begin{itemize}
\item \textbf{Uniqueness:} For each question, there must be a unique and clearly correct answer.
\item \textbf{Concise Language:} The wording of answer should be concise and clear, avoiding complex sentence structures and uncommon vocabulary.
\end{itemize}
\end{itemize}

\item \textbf{Design Distractor Options:}
Design three incorrect distractors based on the question and correct answer. The distractors should also be described clearly, be of consistent length, and be meaningful.
\item \textbf{Option Format Requirements:}
\begin{itemize}
\item \textbf{Consistent Length:} Ensure that four options‘ length are similar, avoid making the correct option easily identifiable due to length differences. 
\item \textbf{Diverse Design:} Design distractors in a varied manner to avoid patterns, avoid consistently employing a singular approach when design.
\item \textbf{Concise Language:} The words of distractor options should be concise and clear, avoiding complex sentence structures and uncommon vocabulary.
\item \textbf{Option significance:} The distractors should be of the same category as the correct answer and should be meaningful in relation to the question stem.

\end{itemize}
\end{enumerate}
\end{methodbox}

\subsubsection{Methods for Designing Incorrect Answer Options}
\begin{methodbox}{Incorrect Option Design Methods}
\begin{itemize}
    \item \textbf{Visual Replacement:} Replace a piece of visual information in the video with information that is similar and incorrect. For example, altering the color or shape of an object.
    \item \textbf{Quantitative Replacement:}  Change the quantity of a detail in the video.
    \item \textbf{Action Replacement:} Describe an action that is similar but different from the actually occurred action.
    \item \textbf{Character Replacement:} Associate the actual event with the wrong character. 
    \item \textbf{Spatial Replacement:} Incorrectly describe the location where an event occurs.
    \item \textbf{Temporal Replacement:} Change the time point of an event in the description.
    \item \textbf{Missing Information:} Create an error by omitting key details in the option. (e.g. leaving out an important action or cause when describing an event)
    \item \textbf{Detail Replacement:} Manufacture an error by exaggerating or minimizing a detail. (e.g. describing “running slowly” as “sprinting quickly”)
    \item \textbf{Sequential Replacement:} Arrange a series of events that occurred in the video in the wrong order.
    \item \textbf{Frequency Replacement:} Repeat the frequency of actions incorrectly.
\end{itemize}
\end{methodbox}

\newtcolorbox[auto counter, number within=section]{purposebox}[2][]{%
  colback=white, 
  colframe=blue!50!black, 
  width=\textwidth,
  arc=2mm, 
  boxrule=0.5mm, 
  title={\normalsize\faCompass\hspace{0.5em}#2},
  breakable, 
  fonttitle=\bfseries\Large, 
  fontupper=\small, 
  drop shadow southeast, 
  #1
}

\newtcolorbox[auto counter, number within=section]{promptbox}[2][]{%
  colback=white, 
  colframe=purple!70!blue!80!black,  
  width=\textwidth,
  arc=2mm, 
  boxrule=0.5mm, 
  title={\normalsize\faInfoCircle\hspace{0.5em}#2},
  breakable,
  fonttitle=\bfseries\Large, 
  fontupper=\small, 
  drop shadow southeast, 
  #1
}
\section{Manual Review Process Details}
The quality inspection of LoneVideoBenchmark comprises two rounds.
\begin{itemize}
    \item \textbf{Round 1:} focuses on standardizing problem structures and correcting elements that do not align with the question.
    \item \textbf{Round 2:} addresses advanced quality requirements to ensure task rigor.
\end{itemize}

\subsection{Round 1 Quality Control}
%\label{Appendix: LLM-Based Screening}

\begin{purposebox}{Purpose}
\begin{itemize}
     \item Ensure the comprehensiveness of the basic structure of the question and the alignment among each elements, including a clear question stem, a correct answer, and meaningful distractors.
\end{itemize}
\end{purposebox}

\begin{promptbox}{Quality Assessment Dimensions}
\begin{enumerate}
    \item \textbf{Question Stem:}
        \begin{itemize}
            \item \textbf{Expression:}Check whether the question stem is coherent and meaningful.
            \item \textbf{Duplicate question stem:} Check whether the question content for the same video are repetitive, if only change the words of question stem consider as the same question.
            \item \textbf{Question Type:} Check whether the question type is correct and whether the question stem corresponds to the question type.
            \item \textbf{Temporal distribution:} Check whether the question stem only focuses on the specific time segments of the video.
        \end{itemize}
    \item \textbf{Options:}
        \begin{itemize}
            \item \textbf{Answer:} Check the correctness of the answer based on the original video.
            \item \textbf{Distractors Type:} Check whether the method of designing incorrect options same to the content of the incorrect options.
            \item \textbf{Distractors content:} Check whether the distractors are meaningful. Distractors should be as same category as the answer and should be meaningful in relation to the question stem.
        \end{itemize}
    \item \textbf{Absolute Time:} Replace question stems that use absolute time references with vague time references.
    \item \textbf{Hierarchy:} Check whether the hierarchy is correctly labeled.
\end{enumerate}
\end{promptbox}

\subsection{Round 2 Quality Control}
\begin{purposebox}{Purpose}
\begin{itemize}
    \item Enhance task validity by verifying multimodal necessity, question stem accurity, and distractor plausibility, thereby preventing evaluation biases caused by design flaws.

\end{itemize}
\end{purposebox}

\begin{promptbox}{Methods}
\begin{enumerate}
    \item \textbf{Information Leakage Detection:} Ensure that the question stem or options do not directly disclose the answer. (e.g. avoid using clothes color to locate a person when asking about the color of his clothes)
    \item \textbf{Commonsense Dependency Screening:} Check whether the content of the question is common knowledge that can be answer directly and whether the question requires prior knowledge to answer. (e.g. "where does the sun rises", "What action did [Trump] take?").
    \item \textbf{Duplicate question types:} Check whether the question types in a video are overly concentrated, and reannotate videos that are overly concentrated.
    \item \textbf{Distractor Optimization:} Redesign meaningless distractors based on video content. Ideally, distractors should correspond to the question categories and create confusion.
    \item \textbf{Video Quality Filtering:} Remove low-quality videos (difficult to describe with precise language) that cannot support effective question design. 
\end{enumerate}
\end{promptbox}

\end{CJK*}
\end{document}